\documentclass[10pt,twocolumn,letterpaper]{article}

\usepackage[pagenumbers]{cvpr} 

\usepackage{booktabs}
\usepackage{array}
\usepackage{multirow}
\usepackage{graphicx}
\usepackage{xcolor}
\usepackage{pifont}

\definecolor{cvprblue}{rgb}{0.21,0.49,0.74}
\usepackage[pagebackref,breaklinks,colorlinks,allcolors=cvprblue]{hyperref}

\def\confName{CVPR}
\def\confYear{2027}

\title{RoboMite: Benchmarking Agentic Robots in Multi-Turn Manipulation with Cross-Task Dependencies}

\author{Shilin Ma$^{1,2}$\quad Mu Lin$^{1}$\quad Chubin Zhang$^{1}$\quad Deyi Zhu$^{1}$\quad Geyi Fan$^{1}$\quad Jingqi Tian$^{1}$\\
Changyuan Wang$^{1}$\quad Xulong Bai$^{1}$\quad Xuantang Xiong$^{2,\dagger}$\quad Yansong Tang$^{1,\dagger}$\\
$^{1}$Tsinghua University\qquad $^{2}$Tencent Robotics X
}

\begin{document}
\maketitle
\begingroup
\renewcommand{\thefootnote}{\fnsymbol{footnote}}
\footnotetext[2]{Corresponding authors.}
\endgroup
\begin{figure*}[t]
    \centering
    \includegraphics[width=\textwidth]{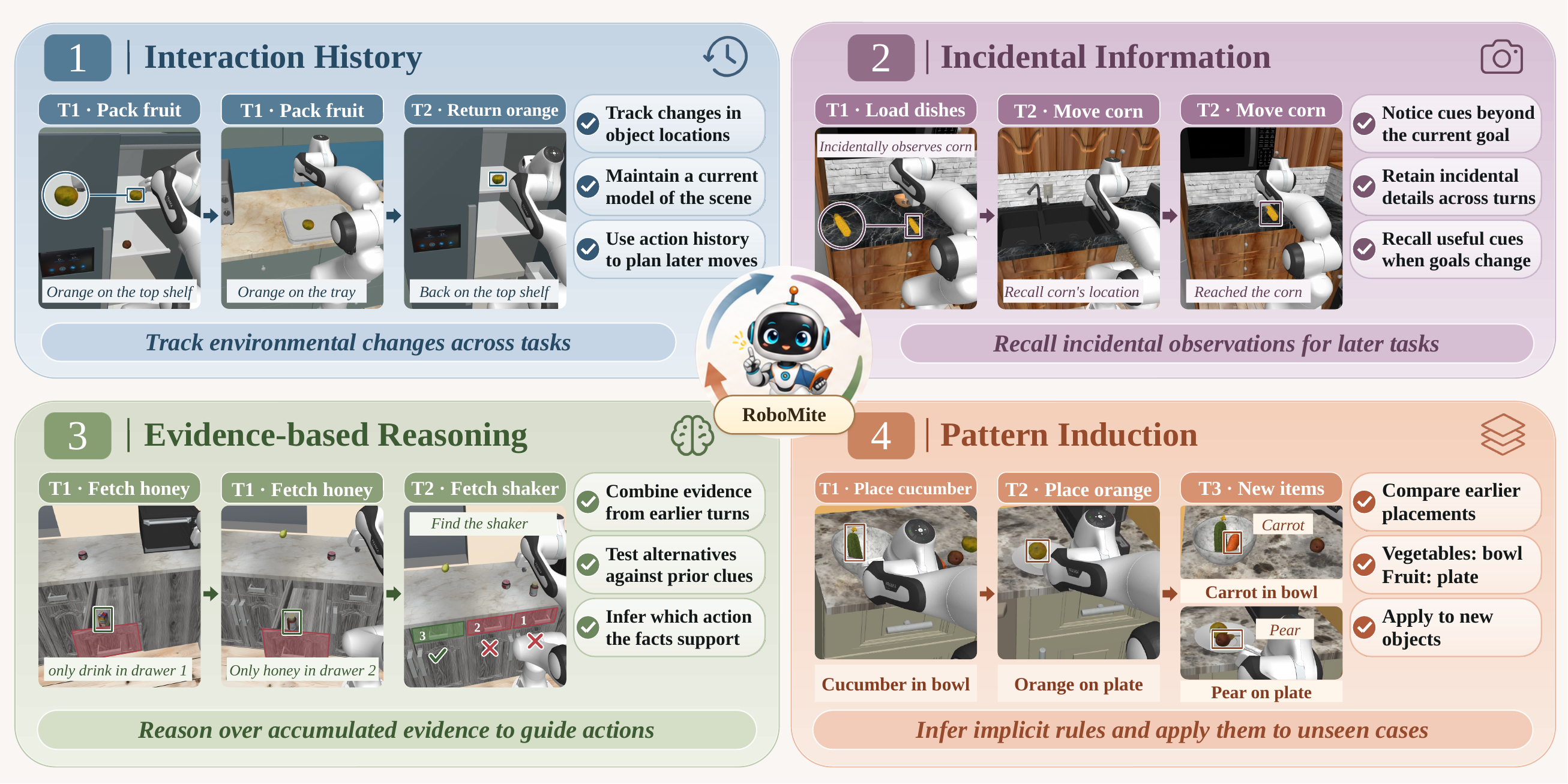}
    \caption{\textbf{Cross-task dependencies in RoboMite.} Interaction-History Dependency requires tracking environmental changes caused by prior actions; Incidental-Information Dependency requires recalling previously observed information unrelated to earlier task objectives; Evidence-Based Reasoning Dependency requires drawing conclusions from accumulated evidence; and Pattern-Induction Dependency requires inferring preferences or behavior patterns from prior interactions and applying them to new tasks. T1, T2, and so on denote Task 1, Task 2, and subsequent tasks in multi-turn evaluation.}
    \label{fig:robomite_dependencies}
    \vspace{-10pt}
\end{figure*}

\begin{abstract}
    Recent advances in agentic robots have demonstrated strong manipulation capabilities by integrating foundation models with low-level control policies. However, existing manipulation benchmarks predominantly evaluate agents through isolated task episodes, leaving their ability to leverage accumulated experience across successive tasks insufficiently explored.  To bridge this gap, we introduce RoboMite, a benchmark for \underline{m}ulti-round \underline{i}nterdependent manipulation \underline{t}ask \underline{e}valuation, where later tasks depend on information acquired or generated during earlier tasks. Specifically, RoboMite encompasses four types of cross-task dependencies: Interaction-History for tracking changes caused by prior actions, Incidental-Information for recalling details incidentally observed during earlier tasks, Evidence-Based Reasoning for determining subsequent actions by integrating evidence, and Pattern-Induction for inferring and applying user preferences. We construct 17 multi-turn task suites and develop an LLM-driven data generation pipeline, yielding 1,700 high-quality demonstrations comprising 16,622 annotated atomic task segments. We further establish a unified evaluation framework spanning agent harnesses, foundation models, and manipulation tools. Experiments demonstrate that structured representations of prior interactions improve full-sequence success by 4.9 percentage points compared with a keyframe-only baseline. Furthermore, while VLA-based manipulation slightly outperforms Code-as-Policy execution, providing privileged object pose information substantially improves the latter's full-sequence success by 21.1 percentage points, highlighting the importance of accurate object grounding.
    \end{abstract}    
\section{Introduction}
\label{sec:introduction}

Recent advances in foundation models have significantly enhanced the capabilities of agentic robots, enabling flexible manipulation across diverse tasks and environments~\cite{jia2026agentaspolicy,zhang2026harnessvla,wang2026sear,si2026pyrua,ma2026activeexploration}. By combining high-level reasoning and planning with low-level execution mechanisms, including vision-language-action (VLA) policies~\cite{kim2024openvla,kim2025openvlaoft,black2025pi05,nvidia2025gr00tn1,zheng2025xvla} and language-model-generated programs~\cite{liang2023codeaspolicies,huang2023voxposer,huang2023instruct2act,fu2026capx}, agentic robots can execute complex tasks, coordinate diverse manipulation skills, and adapt to execution feedback.

Despite these advances, existing manipulation benchmarks predominantly evaluate agents through \textbf{isolated task episodes}~\cite{liu2023libero,chen2025robotwin,james2020rlbench,petkov2026ladder,li2026rmmbench}. However, in real-world applications, agentic robots are often required to execute \textbf{multiple interdependent tasks}, where information acquired during earlier interactions can be valuable for subsequent tasks~\cite{lei2025robomemory,wang2026memoreval}. For example, a household robot may be asked to tidy up toys after cleaning the floor, benefiting from toy locations incidentally observed during cleaning. Successfully handling such cross-task dependencies requires agents not only to execute individual tasks but also to retain, reason over, and leverage accumulated experience across task boundaries. However, existing manipulation benchmarks rarely evaluate such interdependent multi-turn tasks~\cite{liu2023libero,chen2026rmbench,lei2026robomemarena}, leaving a gap between benchmark settings and the demands of real-world applications.

To bridge this gap, we introduce \textbf{RoboMite}, a multi-turn manipulation benchmark designed to systematically evaluate four types of cross-task dependencies: (1) \textbf{Interaction-History Dependency}, which requires tracking environmental changes resulting from previous actions; (2) \textbf{Incidental-Information Dependency}, which requires recalling information incidentally observed during earlier tasks; (3) \textbf{Evidence-Based Reasoning Dependency}, which requires reasoning over accumulated information to infer appropriate actions for subsequent tasks; and (4) \textbf{Pattern-Induction Dependency}, which requires identifying personal preferences and behavior patterns from previous interactions and applying them to new tasks. Figure~\ref{fig:robomite_dependencies} illustrates representative examples of these four dependency types.

Unlike prior long-horizon memory benchmarks~\cite{lei2026robomemarena,dai2026robomme,chen2026rmbench}, RoboMite evaluates not only memory retention but also reasoning over accumulated experience and induction of underlying patterns. More importantly, rather than evaluating isolated episodes, RoboMite adopts a multi-turn setting where task instructions are revealed sequentially and \textbf{future tasks remain unknown}. Agents must therefore preserve potentially useful information without knowing its future relevance, and apply accumulated experience to subsequent tasks. This setting introduces greater challenges in cross-task reasoning and experience utilization, while better reflecting real-world robotic applications.

We manually design 17 multi-turn task suites with explicit cross-task information dependencies and executable success conditions. To enable scalable data collection, we develop an LLM-driven self-improvement pipeline that composes predefined robotic skills into task programs and refines them through simulation feedback. Collected trajectories undergo replay-based verification, human review, and hierarchical annotation. In total, we collect 1,700 high-quality demonstration trajectories comprising 3.43 million control steps and 16K annotated atomic task segments.

Building upon RoboMite, we establish a unified evaluation framework comprising harnesses for cross-task experience representation, foundation models for reasoning and planning, and manipulation tools for physical execution. We evaluate 24 configurations combining two harnesses, four foundation models, and three manipulation tools, covering VLA-based~\cite{team2026xiaomi} and Code-as-Policy (CaP)~\cite{liang2023codeaspolicies,chen2026gap,lu2026aspire} execution. Our experiments show that structured experience improves overall success, model performance gaps vary with manipulation interfaces, and object localization and execution robustness affect multi-turn performance. Our main contributions are summarized as follows: 
\begin{itemize}
    \item We introduce RoboMite, a multi-turn robotic manipulation benchmark featuring 17 task suites and four types of cross-task dependencies, with 1,700 high-quality demonstration trajectories.

    \item We establish a unified evaluation framework spanning agent harnesses, foundation models, and manipulation tools, enabling systematic assessment of cross-task experience utilization, reasoning and planning, and manipulation execution.

    \item We conduct comprehensive experiments across 24 agent configurations, revealing how experience representations, foundation model capabilities, and manipulation strategies jointly influence performance and identifying key bottlenecks in multi-turn robotic manipulation.
\end{itemize}

\section{Related Work}
\label{sec:related_work}

\subsection{Agentic Robots}

Recent agentic robots integrate the perception, reasoning, and planning capabilities of foundation models with robot-control interfaces, enabling flexible coordination of learned policies and executable skills~\cite{liang2023codeaspolicies,zhang2026harnessvla}. Agent as Policy~\cite{jia2026agentaspolicy} places program generation and physical execution under the control of a general-purpose agent. Harness VLA~\cite{zhang2026harnessvla} combines frozen VLA policies with analytic manipulation primitives, while HarnessPAI~\cite{wang2026harnesspai} improves executable control programs through rollout feedback. PyRUA-Lean~\cite{si2026pyrua} further supports conditional execution, local retries, and selective observations to improve efficiency. These advances demonstrate the potential of agentic manipulation systems and motivate evaluating how their reasoning, memory, and execution capabilities interact across distinct but interdependent tasks.

\subsection{Robotic Manipulation Benchmarks}

Robotic manipulation benchmarks evaluate complementary aspects of policy learning, skill composition, and generalization. LIBERO~\cite{liu2023libero} studies knowledge transfer in lifelong robot learning through diverse manipulation task suites. RoboTwin 2.0~\cite{chen2025robotwin} emphasizes robust bimanual manipulation through automated demonstration generation and structured domain randomization. RoboCasa365~\cite{nasiriany2026robocasa365} broadens evaluation to household mobile manipulation across diverse kitchen tasks and environments, supporting multi-task and lifelong learning. CALVIN~\cite{mees2022calvin} evaluates long-horizon language-conditioned manipulation through chained task instructions, testing whether learned skills can be composed and generalized to unseen settings. These benchmarks provide foundations for evaluating manipulation competence, while cross-task information utilization remains a distinct evaluation objective.

Memory-oriented manipulation benchmarks further examine how policies retain and use historical information. RoboMME~\cite{dai2026robomme} evaluates temporal, spatial, object, and procedural memory in history-dependent manipulation. RMBench~\cite{chen2026rmbench} introduces tasks with varying memory complexity to analyze the relationship between policy architecture and memory performance. RoboMemArena~\cite{lei2026robomemarena} extends evaluation to long-horizon tasks with memory-related annotations and paired real-world tasks. RoboMite complements these efforts by explicitly evaluating information dependencies across distinct tasks whose instructions arrive sequentially within a shared environment. Future instructions are withheld, requiring agents to retain observations and interaction outcomes whose relevance may emerge only in later tasks. This setting assesses historical state tracking, incidental-information retrieval, evidence-based reasoning, and preference induction across task boundaries.

\section{RoboMite Benchmark}
\label{sec:robomite}

\begin{table*}[t]
    \centering
    \begingroup
    \newcommand{\rbcheck}{\textcolor{green!65!black}{\ding{51}}}
    \newcommand{\rbcross}{\textcolor{red!85!black}{\ding{55}}}
    \caption{\textbf{Comparison with representative robotic manipulation benchmarks.} \textbf{Long horizon}: average trajectory length exceeding 1,000 control steps; \textbf{multi-turn tasks}: task instructions issued sequentially within a single interaction sequence; \textbf{cross-task dep.}: dependencies between distinct tasks; \textbf{mobile manip.}: mobile manipulation; \textbf{atomic annot.}: fine-grained subtask annotations; \textbf{auto. generation}: automated generation of executable demonstration trajectories; \textbf{memory}: retaining and using historical information; \textbf{preferences \& patterns}: inferring user preferences or behavioral patterns from prior interactions. Green checks indicate explicitly supported features, while red crosses indicate features not explicitly supported.}
    \label{tab:benchmark_comparison}
    \renewcommand{\arraystretch}{1.08}
    \setlength{\tabcolsep}{3pt}
    \small
    \begin{tabular*}{\textwidth}{@{\extracolsep{\fill}}lcccccccc@{}}
        \toprule
        \textbf{Benchmarks} &
        \multicolumn{3}{c}{\textbf{Task Structure}} &
        \multicolumn{3}{c}{\textbf{Benchmark Properties}} &
        \multicolumn{2}{c}{\textbf{Information Demands}} \\
        \cmidrule(lr){2-4} \cmidrule(lr){5-7} \cmidrule(l){8-9}
        & \shortstack{\textbf{Long}\\\textbf{Horizon}}
        & \shortstack{\textbf{Multi-turn}\\\textbf{Tasks}}
        & \shortstack{\textbf{Cross-task}\\\textbf{Dep.}}
        & \shortstack{\textbf{Mobile}\\\textbf{Manip.}}
        & \shortstack{\textbf{Atomic}\\\textbf{Annot.}}
        & \shortstack{\textbf{Auto.}\\\textbf{Generation}}
        & \textbf{Memory}
        & \shortstack{\textbf{Preferences}\\\textbf{\& Patterns}} \\
        \midrule
        \multicolumn{9}{@{}l}{\textit{General Manipulation Benchmarks}} \\
        LIBERO-Long~\cite{liu2023libero}              & \rbcross & \rbcross & \rbcross & \rbcross & \rbcross & \rbcross & \rbcross & \rbcross \\
        CALVIN~\cite{mees2022calvin}                  & \rbcross & \rbcheck & \rbcross & \rbcross & \rbcheck & \rbcross & \rbcross & \rbcross \\
        RoboCasa365~\cite{nasiriany2026robocasa365}  & \rbcross & \rbcross & \rbcross & \rbcheck & \rbcheck & \rbcheck & \rbcross & \rbcross \\
        RoboTwin 2.0~\cite{chen2025robotwin}          & \rbcross & \rbcross & \rbcross & \rbcross & \rbcross & \rbcheck & \rbcross & \rbcross \\
        BEHAVIOR-1K~\cite{li2023behavior1k}           & \rbcheck & \rbcross & \rbcross & \rbcheck & \rbcheck & \rbcross & \rbcross & \rbcross \\
        VLABench~\cite{zhang2025vlabench}             & \rbcross & \rbcross & \rbcross & \rbcross & \rbcheck & \rbcheck & \rbcross & \rbcross \\
        RoboCerebra~\cite{han2025robocerebra}         & \rbcheck & \rbcross & \rbcross & \rbcross & \rbcheck & \rbcross & \rbcheck & \rbcross \\
        RobotAgentBench~\cite{mao2026robotagentbench} & \rbcheck & \rbcross & \rbcross & \rbcheck & \rbcheck & \rbcheck & \rbcheck & \rbcross \\
        \midrule
        \multicolumn{9}{@{}l}{\textit{Memory-oriented Manipulation Benchmarks}} \\
        MIKASA-Robo-VLA~\cite{cherepanov2026mikasarobovla} & \rbcross & \rbcross & \rbcross & \rbcross & \rbcross & \rbcheck & \rbcheck & \rbcross \\
        MEMOBench~\cite{sun2026memobench}             & \rbcross & \rbcross & \rbcross & \rbcross & \rbcheck & \rbcross & \rbcheck & \rbcross \\
        RMBench~\cite{chen2026rmbench}                & \rbcross & \rbcross & \rbcross & \rbcross & \rbcheck & \rbcheck & \rbcheck & \rbcross \\
        RoboMME~\cite{dai2026robomme}                 & \rbcross & \rbcross & \rbcross & \rbcross & \rbcheck & \rbcheck & \rbcheck & \rbcross \\
        RoboMemArena~\cite{lei2026robomemarena}       & \rbcheck & \rbcross & \rbcross & \rbcross & \rbcheck & \rbcheck & \rbcheck & \rbcross \\
        \midrule
        \textbf{RoboMite (Ours)}                      & \rbcheck & \rbcheck & \rbcheck & \rbcheck & \rbcheck & \rbcheck & \rbcheck & \rbcheck \\
        \bottomrule
    \end{tabular*}
    \endgroup
\end{table*}

\subsection{Task Setting}

We consider multi-turn robotic manipulation, where an agentic robot sequentially receives and executes multiple distinct manipulation tasks within a shared environment. A complete multi-turn episode is represented as a temporally ordered task sequence:
\begin{equation}
    \mathcal{T} = (T_1, T_2, \ldots, T_K),
\end{equation}
where $K$ denotes the number of turns. Each turn $T_k$ corresponds to a distinct manipulation task, defined by a natural-language instruction $g_k$ and its resulting observation-action trajectory $\tau_k$:
\begin{equation}
    T_k = (g_k, \tau_k), \qquad
    \tau_k = \{(o_{k,t}, a_{k,t})\}_{t=1}^{H_k},
\end{equation}
where $o_{k,t}$ and $a_{k,t}$ denote the observation and action at execution step $t$, respectively, and $H_k$ is the trajectory length. Task instructions are issued sequentially, without revealing future tasks to the agent.

RoboMite focuses on interdependent multi-turn manipulation, where successful execution of subsequent tasks requires information acquired during preceding tasks. We categorize these cross-task dependencies into four types:

\noindent
\textbf{(1) Interaction-History Dependency (IH)} requires tracking environmental changes resulting from previous actions, such as recalling an object's original location and returning it there after moving it in an earlier task.

\noindent
\textbf{(2) Incidental-Information Dependency (II)} requires recalling information incidentally observed during earlier tasks, including details that may be unrelated to their objectives, such as locating an object previously seen while performing an unrelated task.

\noindent
\textbf{(3) Evidence-Based Reasoning Dependency (ER)} requires reasoning over accumulated information to infer appropriate actions for subsequent tasks, such as searching the third drawer after ruling out the first two.

\noindent
\textbf{(4) Pattern-Induction Dependency (PI)} requires identifying personal preferences and behavior patterns from previous interactions and applying them to new tasks, such as learning a user's utensil-placement preferences and applying them when setting a new plate.

Together, these dependencies evaluate how agentic robots leverage information accumulated across tasks, ranging from historical state tracking and incidental observation retrieval to logical inference and the application of personal preferences and behavior patterns to new tasks. Figure~\ref{fig:robomite_dependencies} illustrates representative examples of these dependencies.

\subsection{Benchmark Construction}
\label{sec:benchmark_construction}

\begin{figure*}[t]
    \centering
    \includegraphics[width=\textwidth]{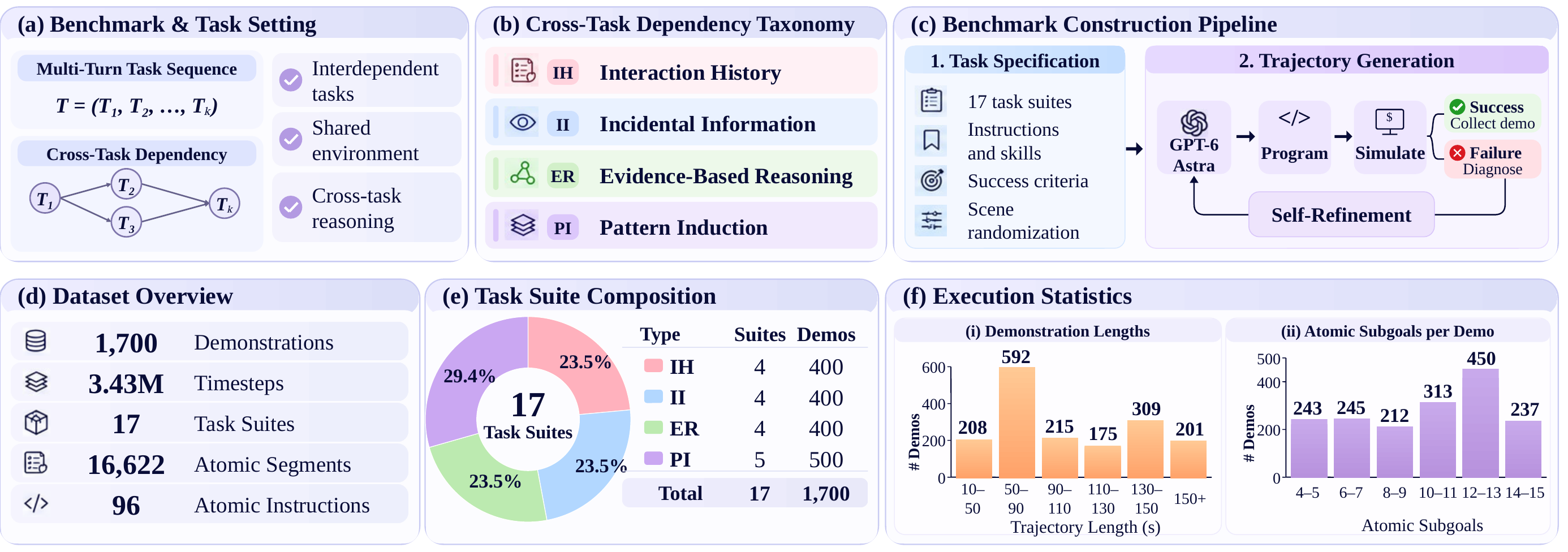}
    \caption{\textbf{RoboMite benchmark overview and dataset statistics.} (a--c) Multi-turn task setting, four types of cross-task dependencies, and the automated trajectory generation pipeline. (d--f) Dataset scale, task suite composition, and distributions of trajectory lengths and atomic subgoals. The dataset contains 1,700 demonstrations across 17 task suites, approximately 3.43 million timesteps, and 16,622 atomic segments with 96 distinct atomic instructions.}
    \label{fig:dataset_statistics}
\end{figure*}

We construct RoboMite in MuJoCo~\cite{todorov2012mujoco}, leveraging simulation assets from RoboCasa365~\cite{nasiriany2026robocasa365} to create kitchen environments for household manipulation tasks with a PandaOmron mobile manipulator. The benchmark construction comprises three stages.

\noindent\textbf{Stage 1. Human-Guided Task Specification.}
We manually design 17 multi-turn task suites with explicit cross-task information dependencies. Each suite specifies task-level language instructions, atomic task descriptions, and executable success conditions. Object positions and scene configurations are randomized under manipulation feasibility constraints. The robot is initialized at a fixed location at the start of each task.

\noindent\textbf{Stage 2. Self-Evolving Automated Trajectory Generation.}
To automate demonstration collection, we employ GPT-6 Astra to generate executable task programs by composing predefined robotic skills, including navigation, grasping, placement, and articulated-object manipulation. Each program specifies a sequence of skill calls to accomplish a given manipulation task. For example, placing an object into a drawer involves opening the drawer, grasping the object, placing it inside, and closing the drawer. The generated programs are executed in simulation under randomized scene configurations, where task success and execution failures are automatically evaluated. Based on this feedback, GPT-6 Astra iteratively refines the programs by adjusting skill sequences, manipulation poses, and recovery strategies. Successful executions are recorded as candidate trajectories, enabling automated data collection across diverse scene configurations. This process enables scalable demonstration collection without requiring manual teleoperation for individual trajectories.

\noindent\textbf{Stage 3. Trajectory Verification and Annotation.}
Candidate trajectories are verified by replaying them in simulation and checking task completion, execution order, and motion quality. Failed or low-quality trajectories are filtered out, followed by manual review to ensure data quality. For each accepted multi-turn trajectory, we associate every task with its corresponding high-level language instruction. Each task is further segmented into atomic tasks, with fine-grained language annotations such as "pick up the target object" and "open the drawer." The resulting demonstrations provide synchronized visual observations, robot states, actions, and hierarchical language annotations. Table~\ref{tab:benchmark_comparison} compares RoboMite with existing manipulation benchmarks.

\subsection{Dataset Statistics}
\label{sec:dataset_statistics}

As illustrated in Fig.~\ref{fig:dataset_statistics}, RoboMite comprises 1,700 high-quality demonstrations across 17 multi-turn task suites, covering four types of cross-task dependencies: Interaction-History (4 suites), Incidental-Information (4), Evidence-Based Reasoning (4), and Pattern-Induction (5). The dataset spans 12 kitchen layouts, 15 scene styles, and 16 layout--style combinations, providing diverse environments for mobile manipulation.

RoboMite contains approximately 3.43 million control steps (47.62 hours at 20 Hz). Each demonstration averages 2,017 steps (100.8 seconds), with 87.8\% exceeding 1,000 steps, highlighting the long-horizon nature of multi-turn manipulation. Furthermore, the dataset provides 16,622 annotated atomic task segments (9.78 per demonstration), covering 96 distinct instructions and six manipulation task types. These statistics demonstrate the benchmark's diversity in task dependencies, temporal complexity, and manipulation requirements.

\subsection{Evaluation Protocol}
\label{sec:evaluation_protocol}

We evaluate complete multi-turn task sequences with sequentially issued instructions in a shared environment, using two complementary metrics: Full-Sequence Success Rate (FSR) and Atomic Completion Rate (ACR).

\noindent\textbf{Full-Sequence Success Rate (FSR).}
FSR measures the proportion of evaluation sequences in which all constituent manipulation tasks are successfully completed. For the $n$-th sequence containing $K_n$ tasks, let $s_{n,k}\in\{0,1\}$ indicate whether the $k$-th task satisfies its success condition.

\noindent\textbf{Atomic Completion Rate (ACR).}
ACR is the ratio of the total number of successfully completed atomic tasks to the total number of atomic tasks across all sequences. Let $M_n$ denote the total number of atomic tasks in sequence $n$, and $z_{n,m}\in\{0,1\}$ indicate whether the $m$-th atomic task is successfully completed. The two metrics are formulated as:
\begin{equation}
\begin{aligned}
\mathrm{FSR}
&= \frac{1}{N}\sum_{n=1}^{N}
\prod_{k=1}^{K_n}s_{n,k},\\[6pt]
\mathrm{ACR}
&= \frac{\sum_{n=1}^{N}\sum_{m=1}^{M_n}z_{n,m}}
{\sum_{n=1}^{N}M_n},
\end{aligned}
\label{eq:evaluation_metrics}
\end{equation}
where $N$ is the total number of evaluated sequences. FSR captures end-to-end reliability across interdependent tasks, while ACR quantifies partial progress even when a sequence is not fully completed. Both metrics are reported overall and separately for each dependency category.

\section{Agentic Robot Manipulation Policies}
\label{sec:agent_policies}

\begin{figure}[t]
    \centering
    \includegraphics[width=\columnwidth]{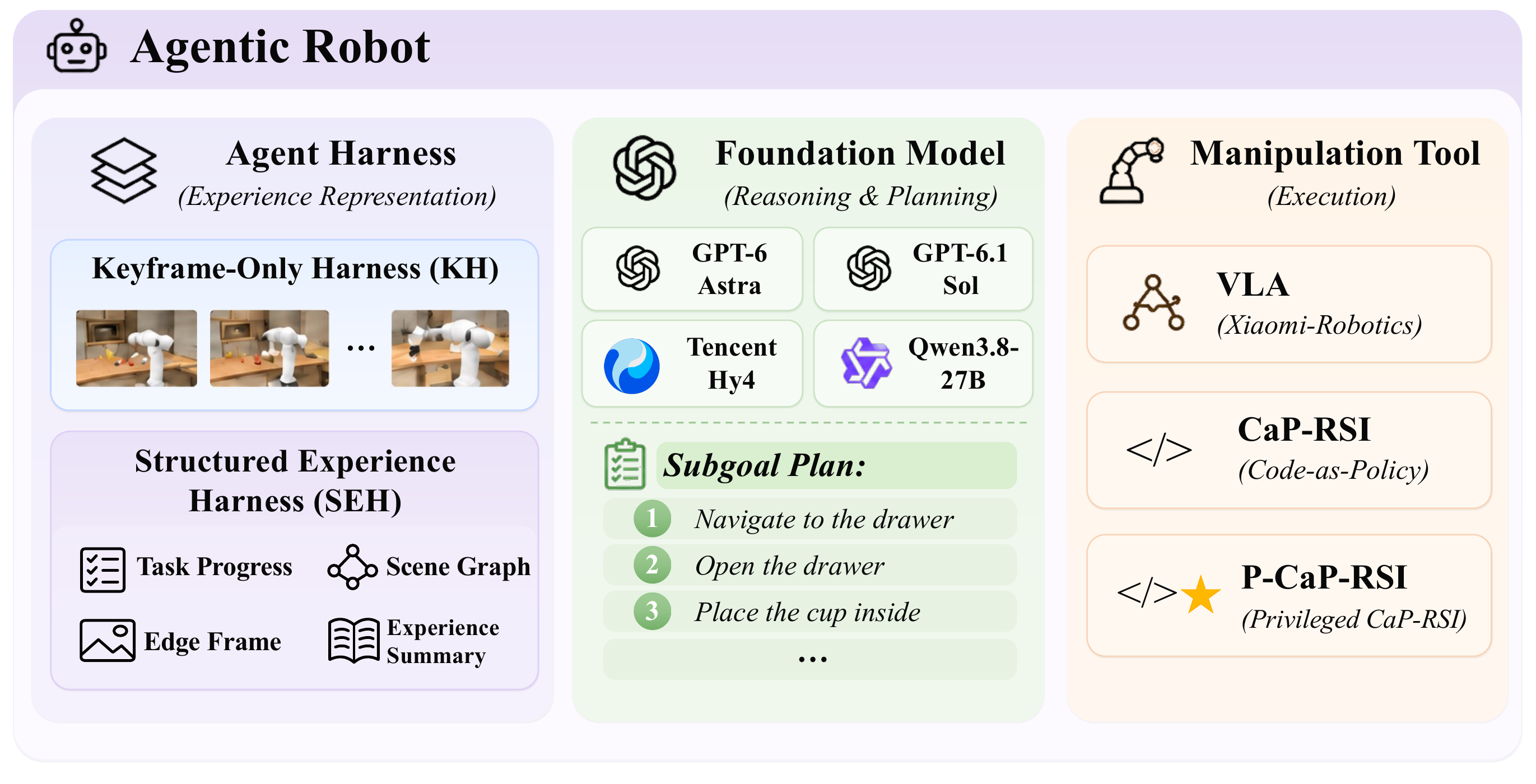}
    \caption{\textbf{Agentic robot evaluation framework.} Agent harnesses retain cross-task experience, foundation models reason over this experience to generate subgoal plans, and manipulation tools execute the resulting plans. We compare keyframe-only and structured experience harnesses, multiple foundation models, and three manipulation strategies: VLA, CaP-RSI, and privileged CaP-RSI.}
    \label{fig:agent_framework}
\end{figure}

Unlike isolated manipulation tasks, multi-turn interdependent manipulation requires agentic robots to coordinate three complementary capabilities: (1) \textbf{cross-task experience representation and retention}, which organizes and preserves historical interactions, incidental observations, and reusable knowledge; (2) \textbf{experience-aware reasoning and planning}, which interprets task instructions, reasons over current observations and available historical experience, and formulates executable subgoal sequences; and (3) \textbf{reliable manipulation execution}, which translates high-level decisions into physical actions. As shown in Fig.~\ref{fig:agent_framework}, to systematically investigate these capabilities, we establish a unified evaluation framework comprising three components: \textit{agent harnesses}, \textit{foundation models}, and \textit{manipulation tools}. By varying their configurations, we examine how different components influence agentic robot performance across the four types of cross-task dependencies.

\subsection{Agent Harnesses}
\label{sec:agent_harnesses}

To investigate how different representations of cross-task experience affect agent performance, we consider two harness configurations that differ in how historical information is organized and retained.

\noindent\textbf{Keyframe-Only Harness (KH).}
This configuration preserves historical information exclusively through keyframe observations collected during preceding tasks. These visual records provide access to past environmental states and interactions without explicitly maintaining structured task progress, scene representations, or summarized experience. It serves as a baseline for evaluating whether visual history alone is sufficient to support cross-task reasoning and planning in multi-turn manipulation settings.

\noindent\textbf{Structured Experience Harness (SEH).}
This configuration maintains cross-task experience through four complementary components: (1) \textit{Task Progress}, which records historical actions and task completion status; (2) \textit{Scene Graph}, which represents objects and their relationships as the environment changes; (3) \textit{Edge Frame}, which associates keyframe observations with scene-graph edges to preserve visual evidence of object relationships and incidental information; and (4) \textit{Experience Summary}, which captures inferred patterns and reusable knowledge from previous tasks. As with keyframes in KH, all information in SEH is selected and organized by the model from observations and interaction history, rather than obtained directly from simulator ground truth. Together, these representations support cross-task reasoning, planning, and adaptation.

\subsection{Foundation Models}
\label{sec:foundation_models}

To investigate the influence of reasoning and planning capabilities, we evaluate four foundation models: \textbf{GPT-6 Astra}, \textbf{GPT-6.1 Sol}, \textbf{Tencent Hy4}, and \textbf{Qwen3.8-27B}. These models serve as the high-level decision-making backbone of agentic robots, interpreting task instructions, reasoning over current observations and available historical experience, and decomposing high-level objectives into atomic subgoals. All models operate through a unified agent interface, enabling systematic comparisons across different cross-task dependencies.

\subsection{Manipulation Tools}
\label{sec:manipulation_tools}

To investigate the influence of low-level execution capabilities and perceptual accuracy, we consider three manipulation strategies that translate atomic subgoals into executable robotic actions.

\noindent\textbf{VLA-based Manipulation.}
We employ Xiaomi-Robotics-1~\cite{team2026xiaomi} as the vision-language-action policy, fine-tuned on 1,700 demonstration trajectories collected in RoboMite. The foundation model decomposes high-level tasks into atomic subgoals, which are subsequently translated into low-level robotic actions by the VLA policy conditioned on visual observations and subgoal instructions.

\noindent\textbf{Code-as-Policy with Recursive Self-Improvement (CaP-RSI).}
Building upon CaP-X~\cite{fu2026capx}, we develop a code-based manipulation framework incorporating the Segment Anything Model 3 (SAM 3)~\cite{carion2026sam3} for object segmentation and Contact-GraspNet~\cite{sundermeyer2021contactgraspnet} for grasp pose estimation. During offline skill evolution, we adopt a recursive self-improvement (RSI) approach using GPT-6 Astra to autonomously generate and refine reusable manipulation skills, including grasping, placement, and articulated-object manipulation. Through iterative code revisions guided by execution feedback, these skills are progressively optimized on training trajectories. During evaluation, the foundation model directly invokes the corresponding pre-established skill interfaces without further modifying implementations.

\noindent\textbf{Privileged CaP-RSI (P-CaP-RSI).}
To isolate the impact of perceptual errors, we introduce a privileged variant of CaP-RSI that shares the same offline-evolved manipulation skills and execution interfaces but replaces perception-based object localization with simulator-provided ground-truth poses of currently visible objects. Hidden object locations remain unavailable, preserving the need for exploration and cross-task experience utilization. This setting allows us to assess the impact of object localization errors on manipulation performance.

\begin{table*}[t]
    \centering
    \caption{\textbf{Performance comparison of agentic robots on RoboMite across four
    cross-task dependency categories.} Each manipulation tool is evaluated with
    two agent harnesses (KH: Keyframe-Only Harness; SEH: Structured Experience Harness)
    and four foundation models. FSR and ACR denote
    Full-Sequence Success Rate and Atomic Completion Rate, respectively.
    Average FSR is weighted by the number of evaluated sequences across categories.
    Average ACR is weighted by the total number of atomic tasks in each category.
    Best results within each manipulation tool are highlighted in bold red (including ties).
    All results are reported in percentages.}
    \label{tab:main_results}
    \small
    \setlength{\tabcolsep}{2.5pt}
    \renewcommand{\arraystretch}{0.9}
    \setlength{\aboverulesep}{1.5pt}
    \setlength{\belowrulesep}{1.5pt}
    \begin{tabular*}{\textwidth}{@{\extracolsep{\fill}}lll*{10}{c}@{}}
    \toprule
    \multirow{2}{*}{\textbf{Tool}}
    & \multirow{2}{*}{\textbf{Harness}}
    & \multirow{2}{*}{\shortstack[l]{\textbf{Foundation}\\\textbf{Model}}}
    & \multicolumn{2}{c}{\shortstack{\textbf{Interaction-}\\\textbf{History}}}
    & \multicolumn{2}{c}{\shortstack{\textbf{Incidental-}\\\textbf{Information}}}
    & \multicolumn{2}{c}{\shortstack{\textbf{Evidence-Based}\\\textbf{Reasoning}}}
    & \multicolumn{2}{c}{\shortstack{\textbf{Pattern-}\\\textbf{Induction}}}
    & \multicolumn{2}{c}{\textbf{Average}} \\
    \cmidrule(lr){4-5}\cmidrule(lr){6-7}\cmidrule(lr){8-9}
    \cmidrule(lr){10-11}\cmidrule(l){12-13}
    & & & FSR & ACR & FSR & ACR & FSR & ACR & FSR & ACR & FSR & ACR \\
    \midrule
    \multirow{8}{*}{VLA} & \multirow{4}{*}{KH} & GPT-6 Astra & 33.5 & 49.7 & \textcolor{red}{\textbf{59.0}} & 78.3 & \textcolor{red}{\textbf{34.0}} & \textcolor{red}{\textbf{50.5}} & 29.6 & 55.0 & 38.5 & 56.8 \\
    & & GPT-6.1 Sol & 28.0 & 52.3 & 51.5 & 76.6 & 33.5 & 46.3 & 30.0 & 56.9 & 35.4 & 56.9 \\
    & & Tencent Hy4 & 24.5 & 48.6 & 44.0 & 63.3 & 25.0 & 36.9 & 16.4 & 44.5 & 26.8 & 47.4 \\
    & & Qwen3.8-27B & 23.5 & 42.5 & 22.0 & 59.3 & 6.0 & 31.9 & 29.2 & 48.6 & 20.7 & 45.2 \\
    \cmidrule(l){3-13}
    & \multirow{4}{*}{SEH} & GPT-6 Astra & \textcolor{red}{\textbf{43.5}} & 53.5 & 58.5 & \textcolor{red}{\textbf{81.7}} & 26.5 & 45.2 & \textcolor{red}{\textbf{50.4}} & 62.6 & \textcolor{red}{\textbf{45.1}} & 59.7 \\
    & & GPT-6.1 Sol & 42.0 & \textcolor{red}{\textbf{57.2}} & 52.5 & 77.3 & 26.0 & 47.4 & 38.8 & \textcolor{red}{\textbf{63.8}} & 39.8 & \textcolor{red}{\textbf{60.8}} \\
    & & Tencent Hy4 & 35.0 & 55.2 & 39.5 & 66.0 & 21.0 & 40.9 & 36.8 & 57.9 & 33.3 & 54.9 \\
    & & Qwen3.8-27B & 33.5 & 54.3 & 33.0 & 58.7 & 11.0 & 33.5 & 23.2 & 47.2 & 25.1 & 48.2 \\
    \midrule
    \multirow{8}{*}{CaP-RSI} & \multirow{4}{*}{KH} & GPT-6 Astra & \textcolor{red}{\textbf{33.0}} & 49.0 & 56.0 & 68.6 & 28.5 & 36.9 & 20.4 & 39.2 & 33.6 & 46.7 \\
    & & GPT-6.1 Sol & 22.5 & \textcolor{red}{\textbf{59.7}} & \textcolor{red}{\textbf{63.0}} & \textcolor{red}{\textbf{72.6}} & 18.5 & 32.6 & 16.8 & 42.6 & 29.4 & 50.6 \\
    & & Tencent Hy4 & 22.0 & 56.0 & 50.5 & 71.0 & 19.0 & 29.9 & 20.4 & 48.2 & 27.5 & 50.5 \\
    & & Qwen3.8-27B & 26.0 & 53.6 & 19.0 & 46.0 & 30.5 & 39.5 & 10.8 & 48.6 & 20.9 & 47.5 \\
    \cmidrule(l){3-13}
    & \multirow{4}{*}{SEH} & GPT-6 Astra & 26.5 & 53.0 & 58.5 & 72.2 & \textcolor{red}{\textbf{44.0}} & \textcolor{red}{\textbf{40.1}} & \textcolor{red}{\textbf{28.8}} & \textcolor{red}{\textbf{53.8}} & \textcolor{red}{\textbf{38.8}} & \textcolor{red}{\textbf{53.9}} \\
    & & GPT-6.1 Sol & 31.0 & 55.7 & 56.5 & 70.9 & 41.0 & 39.3 & 24.4 & 43.3 & 37.4 & 50.9 \\
    & & Tencent Hy4 & 31.0 & 50.4 & 36.0 & 55.3 & 33.0 & 38.5 & 28.4 & 51.3 & 31.9 & 49.0 \\
    & & Qwen3.8-27B & 21.0 & 47.6 & 12.0 & 44.1 & 35.0 & 38.0 & 12.0 & 44.7 & 19.5 & 43.9 \\
    \midrule
    \multirow{8}{*}{P-CaP-RSI} & \multirow{4}{*}{KH} & GPT-6 Astra & 37.0 & 60.6 & 66.0 & 83.0 & \textcolor{red}{\textbf{76.5}} & 74.0 & 44.8 & 65.0 & 55.4 & 69.0 \\
    & & GPT-6.1 Sol & 39.5 & 65.6 & 51.5 & 85.3 & 56.0 & 64.9 & \textcolor{red}{\textbf{60.4}} & 71.3 & 52.4 & 70.8 \\
    & & Tencent Hy4 & 18.0 & 58.6 & 62.0 & 83.0 & 75.5 & 77.3 & 36.8 & 57.3 & 47.4 & 66.7 \\
    & & Qwen3.8-27B & 36.0 & 63.6 & 40.0 & 68.3 & 35.0 & 58.1 & 40.4 & 65.2 & 38.0 & 63.8 \\
    \cmidrule(l){3-13}
    & \multirow{4}{*}{SEH} & GPT-6 Astra & \textcolor{red}{\textbf{55.0}} & 68.5 & \textcolor{red}{\textbf{70.5}} & \textcolor{red}{\textbf{88.7}} & 68.5 & 70.5 & 46.4 & 72.1 & \textcolor{red}{\textbf{59.3}} & \textcolor{red}{\textbf{73.8}} \\
    & & GPT-6.1 Sol & 50.5 & 72.8 & 66.0 & 81.3 & \textcolor{red}{\textbf{76.5}} & \textcolor{red}{\textbf{82.4}} & 34.8 & 60.5 & 55.6 & 72.5 \\
    & & Tencent Hy4 & 48.0 & \textcolor{red}{\textbf{75.7}} & 44.5 & 68.7 & 57.5 & 63.2 & 56.4 & \textcolor{red}{\textbf{72.8}} & 51.9 & 70.7 \\
    & & Qwen3.8-27B & 46.0 & 64.1 & 46.5 & 70.7 & 52.0 & 56.0 & 46.0 & 64.2 & 47.5 & 63.5 \\
    \bottomrule
    \end{tabular*}
\end{table*}

\section{Experiments}
\label{sec:experiments}

We conduct extensive experiments on RoboMite to systematically evaluate agentic robots across three key dimensions: cross-task experience representation, high-level reasoning and planning, and low-level manipulation execution. Beyond comparing overall success rates, we investigate how different agent components affect performance across the four types of cross-task dependencies and analyze the limitations of current agentic robots.

\subsection{Experimental Setup}
\label{sec:experimental_setup}

\noindent\textbf{Agent Configurations.}
We evaluate 24 agentic robot configurations by combining four foundation models, two agent harnesses, and three manipulation tools, as described in Sec.~\ref{sec:agent_policies}. All configurations follow the same evaluation protocol, with task instructions issued sequentially within a shared environment.

\noindent\textbf{Evaluation Settings.}
We evaluate agents on all 17 task suites, covering four Interaction-History (IH), four Incidental-Information (II), four Evidence-Based Reasoning (ER), and five Pattern-Induction (PI) suites. Each suite contains 50 evaluation sequences with randomized initial configurations, using environment seeds disjoint from those used for training. 

\noindent\textbf{Evaluation Metrics.}
Following Sec.~\ref{sec:evaluation_protocol}, we report Full-Sequence Success Rate (FSR) and Atomic Completion Rate (ACR). FSR measures successful completion of entire multi-turn sequences, while ACR measures the fraction of successfully completed atomic tasks pooled across all evaluated sequences. Both metrics are reported for each dependency category and for the benchmark overall.

\subsection{Main Results and Analysis}
\label{sec:main_results}

Table~\ref{tab:main_results} summarizes the performance of all 24 agent configurations across four cross-task dependency categories. We analyze the effects of agent harnesses, foundation models, and manipulation tools in turn.

\subsubsection{Analysis of Agent Harnesses}
\label{sec:harness_analysis}

As shown in Table~\ref{tab:main_results}, SEH achieves higher average FSR than KH on three of the four dependency categories, with improvements of 10.0, 4.5, and 5.9 percentage points on IH, ER, and PI, respectively, while exhibiting comparable performance on II. Overall, SEH improves average FSR from 35.5\% to 40.4\%. These results suggest that structured experience representations are particularly beneficial for tracking environmental changes, reasoning over accumulated evidence, and inferring patterns from previous interactions. In contrast, II tasks rely heavily on incidental visual details, which can also be effectively preserved through keyframe histories when needed for subsequent tasks.

\begin{table}[t]
    \centering
    \caption{\textbf{Offline cross-task dependency evaluation.} GPT-6.1 Sol
    is evaluated using saved trajectory videos, without executing actions.
    SEH and KH denote the Structured Experience Harness and
    Keyframe-Only Harness, respectively. Entries report success rates (\%). Overall results are weighted
    by the number of decision points in each dependency category.}
    \label{tab:offline_dependency}
    \small
    \setlength{\tabcolsep}{3pt}
    \renewcommand{\arraystretch}{1.15}
    \begin{tabular*}{\columnwidth}{@{\extracolsep{\fill}}lccccc@{}}
        \toprule
        \textbf{Harness} & \textbf{IH} & \textbf{II} & \textbf{ER} & \textbf{PI} & \textbf{Overall} \\
        \midrule
        None & 3.0 & 0.0 & 10.0 & 18.0 & 5.3 \\
        SEH & \textbf{92.0} & \textbf{75.0} & \textbf{80.0} & \textbf{98.0} & \textbf{86.3} \\
        KH & 64.0 & 72.5 & 77.5 & 62.0 & 69.0 \\
        \bottomrule
    \end{tabular*}
\end{table}

To further isolate cross-task experience utilization from manipulation execution, we conduct an offline dependency evaluation using recorded robot trajectories. GPT-6.1-Sol processes historical trajectories under each harness configuration and is evaluated through question answering at history-dependent decision points, without executing robotic actions. As shown in Table~\ref{tab:offline_dependency}, SEH achieves an overall accuracy of 86.3\%, outperforming KH by 17.3 percentage points. The largest improvements occur on IH and PI tasks, reaching 28.0 and 36.0 points, respectively. In contrast, the no-history baseline, which evaluates each task independently without access to previous interactions, achieves only 5.3\% accuracy, validating the importance of cross-task dependencies in RoboMite. These results suggest that structured experience representations can benefit cross-task reasoning in this evaluation setting.

\begin{table}[t]
    \centering
    \caption{\textbf{Sequence success rates (\%) under oracle atomic instructions.} Pose Reset restores the robot to the corresponding reference pose after each successfully completed atomic task. A sequence succeeds only if all constituent tasks are completed.}
    \label{tab:atomic_execution}
    \small
    \setlength{\tabcolsep}{3pt}
    \renewcommand{\arraystretch}{1.1}
    \begin{tabular*}{\columnwidth}{@{\extracolsep{\fill}}llccc@{}}
        \toprule
        \textbf{Method} & \textbf{Execution} & \textbf{1-Atomic} & \textbf{2-Atomic} & \textbf{3-Atomic} \\
        \midrule
        \multirow{2}{*}{VLA}
        & Free-running & \textbf{92.2} & 84.0 & 64.4 \\
        & Pose Reset & \textbf{92.2} & \textbf{86.7} & \textbf{75.6} \\
        \midrule
        \multirow{2}{*}{CaP-RSI}
        & Free-running & 81.9 & 70.7 & 61.4 \\
        & Pose Reset & 81.9 & 70.7 & 62.8 \\
        \bottomrule
    \end{tabular*}
\end{table}

\subsubsection{Analysis of Foundation Models}
\label{sec:foundation_model_analysis}
The performance gap between foundation models depends on the manipulation interface. Averaging overall ACR across the two harnesses in Table~\ref{tab:main_results}, GPT-6 Astra outperforms Qwen3.8-27B by 11.6 percentage points with VLA, achieving 58.3\% versus 46.7\%. This gap narrows to 4.6 points with CaP-RSI and 7.8 points with P-CaP-RSI. Both code-based settings thus reduce the model gap relative to VLA under the evaluated harness configurations.

One possible explanation is that structured skill interfaces reduce dependence on foundation model capabilities. VLA relies on natural-language subgoals, whereas CaP-RSI and P-CaP-RSI expose pre-established skills through parameterized interfaces, reducing ambiguity and allowing models to focus on skill selection and composition. These findings highlight the importance of jointly considering model selection and manipulation interfaces when designing agentic robots under different resource constraints.

\subsubsection{Analysis of Manipulation Tools}
\label{sec:manipulation_tool_analysis}

Manipulation strategies substantially influence end-to-end task success. As shown in Table~\ref{tab:main_results}, VLA achieves a higher average FSR than CaP-RSI (33.1\% versus 29.9\%), indicating its advantage in overall manipulation performance. More importantly, introducing privileged object poses into CaP-RSI leads to a substantial performance improvement. P-CaP-RSI achieves an average FSR of 50.9\%, outperforming CaP-RSI by 21.1 percentage points and VLA by 17.9 percentage points across the evaluated agent configurations.

This improvement reveals the substantial impact of object localization accuracy on robotic manipulation. With accurate object poses, the code-based manipulation framework achieves considerably stronger performance, even surpassing the VLA-based approach. These findings highlight the potential of code-based manipulation and identify accurate object grounding and pose estimation as important directions for improving agentic robot performance in multi-turn manipulation tasks.

To isolate manipulation execution from high-level task decomposition, we provide GPT-6.1-Sol with oracle atomic instructions and evaluate VLA and CaP-RSI on sequences containing one to three atomic tasks. We compare free-running execution with a reference-pose reset setting, where only the robot pose is restored to the corresponding successful demonstration pose between completed atomic tasks, while the environment state remains unchanged. As shown in Table~\ref{tab:atomic_execution}, both methods exhibit decreasing success rates as sequence length increases. Notably, reference-pose reset improves VLA success from 64.4\% to 75.6\% on three-atomic sequences, a gain of 11.2 percentage points, compared with only 1.4 points for CaP-RSI. These results suggest that VLA is more sensitive to accumulated robot-state deviations during continuous execution, while CaP-RSI exhibits greater robustness to intermediate pose variations.

\section{Conclusion}
\label{sec:conclusion}

In this work, we introduce RoboMite, a comprehensive benchmark for evaluating agentic robots on multi-turn interdependent manipulation tasks within a shared environment. RoboMite encompasses four types of cross-task dependencies and provides 17 task suites with 1,700 high-quality demonstrations generated through an LLM-driven self-evolving pipeline. We further establish a systematic evaluation framework spanning agent harnesses, foundation models, and manipulation tools. Extensive experiments reveal that structured experience and visual keyframe histories offer complementary advantages across different dependency types, while the impact of foundation model capabilities is strongly influenced by manipulation interfaces. Moreover, the substantial improvements achieved with privileged object poses highlight accurate object localization as a critical factor in code-based manipulation. These findings demonstrate that effective multi-turn manipulation requires the integration of cross-task experience, high-level reasoning, and reliable low-level execution. RoboMite provides a comprehensive testbed for understanding these capabilities and advancing more intelligent, experience-aware agentic robots in multi-turn household manipulation settings.


{
    \small
    \bibliographystyle{ieeenat_fullname}
    \bibliography{main}
}



\end{document}